\documentclass[journal]{IEEEtran}
\usepackage{amsmath}
\usepackage{mathtools}
\usepackage[mathscr]{euscript}
\usepackage[noend,ruled,vlined]{algorithm2e}
\usepackage[latin1]{inputenc}
\usepackage{graphicx}
\usepackage{multirow}
\usepackage{array}
\usepackage{subfiles}
\usepackage{url}
\ifCLASSINFOpdf
\else
\fi

\begin{document}
\title{Adaptive Convolution for Semantic Role Labeling}

\author{Kashif~Munir,
        Hai~Zhao,
        and~Zuchao~Li% <-this % stops a space
\thanks{
This paper was partially supported by National Key Research and Development Program of China (No. 2017YFB0304100), Key Projects of National Natural Science Foundation of China (U1836222 and 61733011), Huawei-SJTU long term AI project, Cutting-edge Machine Reading Comprehension and Language Model (Corresponding author: Hai Zhao).
 }

\thanks{Kashif Munir, Hai Zhao, Zuchao Li are with the Department of Computer Science and Engineering, Shanghai Jiao Tong University, and also with Key Laboratory of Shanghai Education Commission for Intelligent Interaction and Cognitive Engineering, Shanghai Jiao Tong University, and also with MoE Key Lab of Artificial Intelligence, AI Institute, Shanghai Jiao Tong University. (Emails: kashifmunir92@sjtu.edu.cn; zhaohai@cs.sjtu.edu.cn; charlee@sjtu.edu.cn)
}% <-this % stops a space
}


\markboth{IEEE/ACM TRANSACTIONS ON AUDIO, SPEECH, AND LANGUAGE PROCESSING}%
{Shell \MakeLowercase{\textit{et al.}}: Bare Demo of IEEEtran.cls for IEEE Journals}



\maketitle

\begin{abstract}
Semantic role labeling (SRL) aims at elaborating the meaning of a sentence by forming a predicate-argument structure. Recent researches depicted that the effective use of syntax can improve SRL performance. However, syntax is a complicated linguistic clue and is hard to be effectively applied in a downstream task like SRL. This work effectively encodes syntax using adaptive convolution which endows strong flexibility to existing convolutional networks. The existing CNNs may help in encoding a complicated structure like syntax for SRL, but it still has shortcomings. Contrary to traditional convolutional networks that use same filters for different inputs, adaptive convolution uses adaptively generated filters conditioned on syntactically-informed inputs. We achieve this with the integration of a filter generation network which generates the input specific filters. This helps the model to focus on important syntactic features present inside the input, thus enlarging the gap between syntax-aware and syntax-agnostic SRL systems. We further study a hashing technique to compress the size of the filter generation network for SRL in terms of trainable parameters. Experiments on CoNLL-2009 dataset confirm that the proposed model substantially outperforms most previous SRL systems for both English and Chinese languages.
\end{abstract}


% \fi
\IEEEpeerreviewmaketitle

\section{Introduction}

\begin{figure}[t]
\centering
  \includegraphics[width=0.4\textwidth]{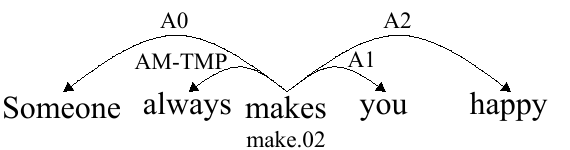} 
  \caption{Semantic role labeling example.}
  \label{sent}
\end{figure}

\IEEEPARstart{S}{emantic} role labeling (SRL), also known as shallow semantic parsing, conveys the meaning of a sentence by forming a predicate-argument structure for each predicate in a sentence, which is generally described as the answer to the question "\textit{Who did what to whom, where and when?}". The relation between a specific predicate and its argument provides an extra layer of abstraction beyond syntactic dependencies (\textit{subject} and \textit{object}) \cite{gildea2002automatic}, such that the labels are insensitive to syntactic alternations and can also be applied to nominal predicates. Given a sentence in Figure \ref{sent}, SRL pipeline framework consists of 4 subtasks, including predicate identification (\textit{makes}), predicate disambiguation (\textit{make.02}), arguments identification (\textit{Someone}) and arguments classification (\textit{Someone} is \textit{A0} for predicate \textit{makes}). SRL is a core task of natural language processing (NLP) having wide range of applications such as neural machine
translation \cite{shi2016knowledge}, information extraction \cite{surdeanu2003using}, question answering \cite{berant2013semantic,yih2016value}, emotion recognition from text \cite{wu2006emotion}, document summarization \cite{yan2014srrank} etc.

Semantic role labeling can be categorized into two categories, span and dependency. Both types of SRL are useful for formal semantic representations but dependency based SRL is better for the convenience and effectiveness of semantic machine learning. Johansson and Nugues \cite{johansson2008dependency} concluded that the best dependency based SRL system outperforms the best span based SRL system through gold syntactic structure transformation. The same conclusion was also verified by Li \textit{et al.} \cite{li2019dependency} through a solid empirical verification. Furthermore, since 2008, dependency based SRL has been more studied as compared to span based SRL.
With this motivation, we focus on dependency based SRL, which is mainly popularized by CoNLL-2008 and CoNLL-2009 shared tasks \cite{surdeanu2008conll,hajivc2009conll}. 

The traditional approaches to SRL focus on feature engineering which struggles in apprehending discriminative information \cite{pradhan2005semantic,zhao2009multilingual} while neural networks are proficient enough to extract features automatically \cite{bai2018deep,zhang2018one}. Specifically, since large scale empirical verification of Punyakanok \textit{et al.} \cite{punyakanok2008importance}, syntactic information has been proven to be extremely beneficial for SRL task. Later works \cite{zhou2015end,he2017deep,marcheggiani2017simple} achieve satisfactory performance for SRL with syntax-agnostic models which creates conflict with the long-held belief that syntax is essential for high-performance SRL \cite{gildea2002necessity}. 
The study of Li \textit{et al.} \cite{li2018unified} shows that the empirical results from neural models on the less importance of syntax indicate a potential challenge and despite the satisfactory performance of syntax-agnostic SRL systems, the reasons behind the absence of syntax in these models are three-fold. First, the effective incorporation of syntax in neural SRL models is quite challenging as compared to traditional approaches. Second, neural SRL models may cover partial syntactic clues more or less. Third, syntax has always been a complicated formalism in linguistics and its not easy to encode syntax for later usage.
%Despite the satisfactory performance of syntax-agnostic SRL models, the reasons behind the absence of syntax in these models are two-fold. First, the effective incorporation of syntax information in neural SRL models is quite challenging. Second, the unreliability of syntactic parsers on account of the risk of erroneous syntactic input may lead to error proliferation. This has been proven by Li \textit{et al.} \cite{li2018unified} through a strong empirical verification. They show that the effective method of syntax incorporation and the high quality of syntax can promote SRL performance.%
Recently, many works have been published for the effective incorporation of syntax in SRL systems. Qian \textit{et al.} \cite{qian2017syntax} propose syntax aware long short term memory (LSTM) to directly model complex syntax parsing information in an architecture engineering way to enhance SRL performance. Similarly, Marcheggiani and Titov \cite{marcheggiani2017encoding} present a syntactic graph convolutional neural network (GCNN) based model for SRL to further enhance the performance. Given the experimental facts that syntax can alleviate SRL performance if incorporated effectively in the neural model, we seek to effectively model complex dependency parsing information in a neural model.

Furthermore, neural networks involving convolution neural networks (CNNs) have shown remarkable achievements in different fields of NLP \cite{kim2014convolutional,johnson2017deep,wang2018densely,cakir2017convolutional,abdel2014convolutional}. The driving force behind CNNs is the use of a convolution operation to screen the local information present in the inputs (either directly from the text or from intermediate hidden states of neural networks) by using a set of filters. 
Convolutional filters are like a pool of questions that ask for the intensity of particular patterns in the inputs and the convolution operation helps in retrieving the answers from the inputs to the questions. However, if the pool of questions is limited to a particular concept related to inputs, the convolution operation will be able to provide more concentrated answers related to questions.
Contrarily, typical CNN architectures use the same set of filters under all circumstances \cite{kim2014convolutional,wang2018densely}, which may stymie CNNs from leveraging the information from the intermediate hidden states and focus the concentration on disentangling uncertainty.

Motivated by this, we present an adaptive convolution for SRL which allows the network to utilize the syntax information they have in the inputs. The adaptive convolution uses the dynamically generated filters (questions) conditioned on inputs.
 We first encode the sentence using BiLSTM and Tree-Structured LSTM \cite{li2018unified,tai2015improved} to model the syntactic information for SRL and then encoder output is fed into a filter generation network, a carefully designed modular network, which generates filters conditioned on syntactically-informed inputs for convolution operation \cite{choi2019adaptive}. The generated filters reflect the syntax information present in the inputs and allow the model to focus on important informative features encoded by BiLSTM and Tree-LSTM encoders. The filter generation network can be easily applied to existing CNN architectures. We further investigate a hashing technique that helps in the compression of the filter generating network to allow the adaptive convolution operation without a considerable increase in the number of parameters. Our major contributions are:

\noindent $\bullet$ A neural framework for SRL which effectively integrates the syntactic information of text.

\noindent $\bullet$ The integration of adaptive convolution in SRL model which helps the model to focus on important informative features encoded by LSTM and Tree-LSTM, and at the same time gives stronger flexibility to existing CNNs.

\noindent $\bullet$ The detailed study of a hashing technique to apply adaptive convolution without a considerable increase in the number of trainable parameters.

\noindent $\bullet$ The proposed model outperforms most previous SRL approaches on CoNLL-2009 English and Chinese datasets.

\begin{figure*}[ht]
\centering
\includegraphics[width=0.7\textwidth]{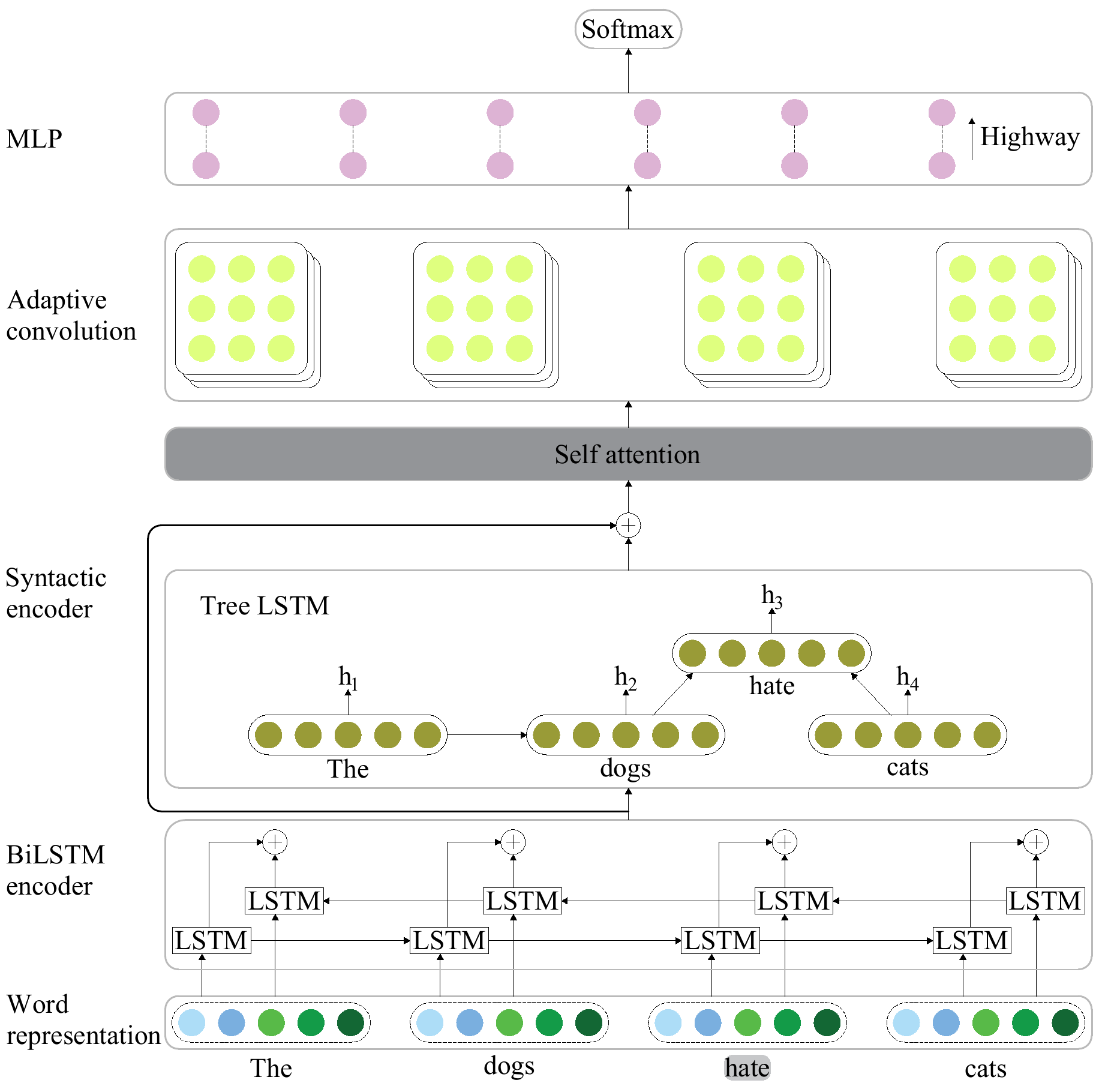}
\caption{Proposed SRL framework.}
\label{framework}
\end{figure*}

\section{Related Work}
Semantic role labeling was pioneered by Gildea and Jurafsky \cite{gildea2002automatic}. In early days of SRL research, a substantial attention has been paid to featured engineering \cite{pradhan2005semantic,zhao2008parsing,zhao2009semantic,zhao2009multilingual2,zhao2009multilingual,li2009improving,bjorkelund2009multilingual,zhao2013integrative}. Pradhan \textit{et al.} \cite{pradhan2005semantic} deploy the SVM classifier and combine features from different syntactic parses, while Zhao \textit{et al.} \cite{zhao2009multilingual} use sets of language-specific features for SRL task. Li \textit{et al.} \cite{li2009improving} integrate features driven from verbal SRL architecture. Bj{\"o}rkelund \textit{et al.} \cite{bjorkelund2009multilingual} propose a beam search in the first stage of their system to label arguments, reranker in the second stage and then combine these scores in the third stage to label arguments for each predicate.

Yang and Zong \cite{yang2016learning} learn generalized feature vectors for arguments with a strong intuition that arguments occurring in the same syntactic positions bear the same syntactic roles. Che \textit{et al.} \cite{che2008using} use a hybrid convolution tree kernel to learn link feature between argument and predicate and syntactic structure features to perform SRL task. Li and Zhou \cite{li2011unified} present a unified framework for SRL for verbal and nominal predicates. Yang \textit{et al.} \cite{yang2016bilingual} use Bi-directional Projection (BDP) method to perform bilingual semantic role labeling.

With the recent success of neural networks \cite{zhang2016probabilistic,cai2016neural,qin2016implicit,wang2016connecting,wang2016bilingual,zhang2018subword,li2018seq2seq,huang2018moon,do2015domain,sundermeyer2015feedforward}, a number of neural network based SRL systems have been proposed \cite{guan2019semantic,he2019syntaxx,li2020high,zhou2019parsing,li2020memory}. Foland and Martin \cite{foland2015dependency} use a convolutional and time-domain neural network to develop a semantic role labeler. FitzGerald \textit{et al.} \cite{fitzgerald2015semantic} present a neural network to jointly embed arguments and their semantic roles, akin to the work \cite{lei2015high} which presents a tensor based approach to induce compact feature representation of the words and their corresponding relations.

Recently, many researchers proposed syntax agnostic models for SRL  \cite{zhou2015end,he2017deep,tan2018deep,cai2018full,li2019dependency} and achieve favorable results without using syntax. Cai \textit{et al} \cite{cai2018full} use a biaffine attention model to propose a full end-to-end syntax agnostic model for SRL. While researchers have been able to produce satisfactory results without syntax, many efforts have been made to effectively integrate syntax in SRL systems. Roth and Lapata \cite{roth2016neural} modeled the syntactic information through dependency path embeddings to achieve notable success. Marcheggiani and Titov \cite{marcheggiani2017encoding} deployed a graph convolutional neural network, while Qian \textit{et al.} \cite{qian2017syntax} used SA-LSTM to encode syntactic information in sentences.  Li \textit{et al} \cite{li2018unified} presented various ways of deploying syntactic information and concluded that the effective integration of syntax can boost SRL performance.

In this work, we follow Li \textit{et al} \cite{li2018unified} to integrate syntax information by using a modified version of Tree LSTM. Owing to the recent success of CNNs in NLP \cite{kim2014convolutional,zhang2015character,johnson2017deep}, we integrate adaptive convolution via a filter generation network in our SRL model. The ability of the filter generation network to produce filters conditioned on inputs allows the model to extract important syntactic features encoded by BiLSTM and Tree-LSTM encoder. We further study the effect of a hashing technique on the compression of a filter generation network in terms of trainable parameters.

\section{Methodology}\label{methodology}
Figure \ref{framework} shows the complete architecture for SRL. Since predicates are already identified in CoNLL-2009 shared task, we focus on the identification and labeling of arguments which can be defined as a sequence tagging problem. Our SRL consists of two main modules: 1) Sentence encoder. 2) Filter generation network. In section \ref{sen-encoder}, we explain the encoder, in section \ref{filtergen}, how to generate filters with a filter generation network and in section \ref{adaptiveconvolution}, how to use generated filters for adaptive convolution.

\subsection{Sentence Encoder}
\label{sen-encoder}
\textbf{Word representation:} Following the previous convention \cite{marcheggiani2017encoding}, we consider predicate-specific word representations for a given sentence and a known predicate. Each word representation $x_i$ is formed by a concatenation of several features: 1) a randomly initialized embedding $x_i^r$. 2) a randomly initialized lemma embedding $x_i^l$. 3) a pre-trained embedding $x_i^p$. 4) a POS tag embedding $x_i^{POS}$. 5) a predicate related information feature $x_i^f$ which is typically a flag \{1,0\} indicating if a particular word is a predicate or not. 6) Embedding from language model ELMO \cite{peters2018deep}. The final word representation will be: $x_i = [x_i^r, x_i^l, x_i^p, x_i^{POS}, x_i^f, {ELMo}_i]$.

\bigbreak
\noindent \textbf{BiLSTM encoder:} Given an input sequence $x = (x_1, x_2, ..., x_m)$ where $m$ is the length of a sequence, we apply bidirectional long short term memory (BiLSTM) \cite{hochreiter1997long} to encode a sequential input. At each time step $t$, we concatenate two hidden states from BiLSTM to get $v_t$.

\begin{gather}
\label{bilstm}
\allowdisplaybreaks
\overleftarrow{v_t} = \overleftarrow{LSTM}(x_t, \overleftarrow{v_{t-1}}),\quad
\overrightarrow{v_t} = \overrightarrow{LSTM}(x_t, \overrightarrow{v_{t+1}})\nonumber\\
 v_t = [\overleftarrow{v_t};\overrightarrow{v_t}]
\end{gather}

\noindent The resulting $v_t$ becomes an input to the next BiLSTM layer. We stack four layers of BiLSTM. 

\noindent \textbf{Syntactic encoder:} For effective integration of syntax in the model, we follow Li \textit{et al.} \cite{li2018unified} and integrate Tree LSTM in our model. Tree LSTM is an extended version of standard LSTM and focuses on modeling tree-structured topologies. This Tree LSTM is an adaption from original Child-Sum Tree LSTM, in which a single forget gate is assigned to each child unit. It takes arbitrarily many child units into account and utilizes them to compose input vectors and hidden states at each time step. For a given syntactic tree, if $n_k$ denotes a current node, $C(k)$ a set of its children and $L(k,.)$ a set of dependency relations between $n_k$ and those nodes having a connection with $n_k$, the Tree LSTM formulation is as follows:
\begin{gather} \label{tree-eq1}
\allowdisplaybreaks
r_g^{k,j} = \sigma (W^{(r)}v_k + U^{(r)}h_j + b^{L(k,j)}), \nonumber\\
\tilde{h}_k = \sum_{j \in C(k)} r_g^{k,j} \odot h_j
\end{gather}
\begin{gather} 
\allowdisplaybreaks
i_g = \sigma (W^{(i)}v_k + U^{(i)}\tilde{h}_k + b^{(i)}), \nonumber\\
f_g^{(k,j)} = \sigma (W^{(f)}v_k + U^{(f)}h_j + b^{(f)}), \nonumber\\
o_g = \sigma (W^{(o)}v_k + U^{(o)}\tilde{h}_k + b^{(o)}), \nonumber\\
\allowdisplaybreaks
u = \tanh (W^{(u)}v_k + U^{(u)}\tilde{h}_k + b^{(u)}), \nonumber\\
c_k = i_g \odot u + \sum_{j \in C(k)} f_g^{(k,j)} \odot c_j, \nonumber\\
h_k = o_g \odot \tanh (c_k)
\vspace{-1em}
\end{gather}

where $h_j$ represents the hidden state of the $j$-th child node, $c_k$ represents a memory cell for the head node $k$ and $b^{L(k,j)}$ is a bias term related to relation label. The output of Tree LSTM $h_k$ is fed into a filter generation network.

\subsection{Filter generation network}
\label{filtergen}
The filter generation network takes the output of syntactic encoder $H = [h_1, h_2, ..., h_m]$ as the input, where $h_k \in {\rm I\!R}^{d}$ is a vector of dimension $d$ at the $k^{th}$ position in the input. The first convolution block takes syntactic encoder's output as input and produces an output with a dimension equal to the number of filters used in that particular convolution block, which becomes an input to the next convolution block and so on. The output of the filter generation network is $n$ number of convolution filters $\mathscr{F}= [f_1, f_2, ..., f_n]$ for each $h_k$, where $f_i \in {\rm I\!R}^{s.d}$ and $s$ being the filter size.

The filter generation network produces filters in two steps: context vector generation and filter generation. The context vectors are generated with the help of a self-attention mechanism and then the filters are generated adaptively from these context vectors.
\bigbreak
\noindent \textbf{Context vectors generation: } Before generating filters, each $h_k$ is self-attended using a special case of self-attention \cite{lin2017structured}. 
\begin{gather}
\label{attention}
C = a_j.h_j \;\;\;s.t \;\;\;j \in (1,m), \nonumber\\
a_j = \frac{\exp q^\top h_j}{\sum_{k=1}^m \exp(q^\top h_k)}
\end{gather}
\noindent where q represents a query vector and is trainable. $C$ contains a context vector $c_k$ for each $h_k$ and now trainable filters can be generated for each context vector. 

\bigbreak
\noindent \textbf{Filter generation: } Once we obtain context vectors by attending the hidden states of syntactic encoder, the filters $\mathscr{F}$ are generated by a function of $C$:
\begin{equation}
    \mathscr{F} = \textbf{f}(C)
\end{equation}

\noindent To train the existing CNNs after the addition of the filter generation network in an end to end fashion we need an architecture whose gradients are differentiable and can be back-propagated. For this purpose, we first use a fully connected layer and then generate filters by using two approaches: \textit{full generation} and \textit{hashed generation}. All the filters in $\mathscr{F}$ are generated in the same manner, so here we will explain the procedure for one filter $f_i$ for one context vector $c_k$. 

\bigbreak
\noindent \textit{Full generation:} A simple way is to use the output of a fully connected layer as a filter. The layer takes $c_k$ as input and generates a filter $f_i$ as follows:
\begin{equation}
   f_i = W_i c_k 
\end{equation}
\noindent where $W_i \in {\rm I\!R}^{(s.d) \times g}$ is a weight matrix for the generation of the $i$-th filter. $g$ is the dimension of a context vector $c_k$.

\begin{figure}[t]
\centering
\includegraphics[width=\linewidth]{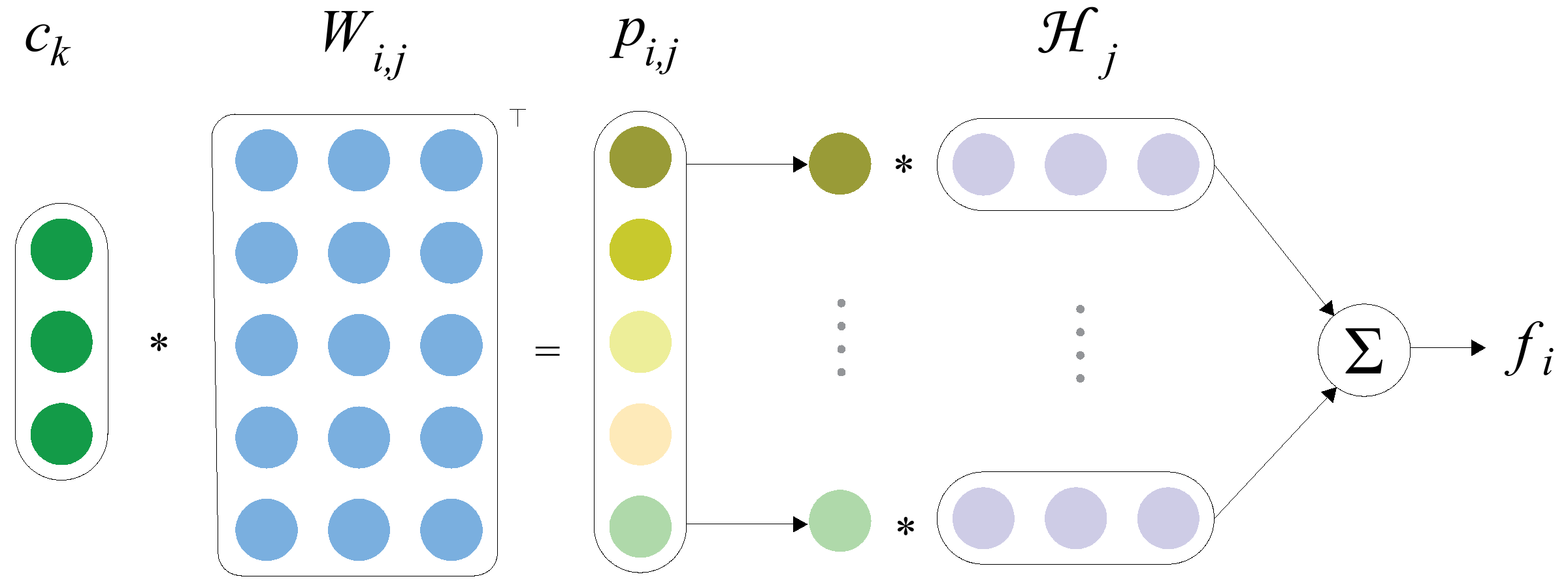}
\caption{Procedure for generating each filter $f_i$. $\mathscr{H}_j$ outputs the component filter by looking up the shared pool $E$ according to the bucket index produced by $D_j$.}
\label{hashgenerationfig}
\end{figure}

\bigbreak
\noindent \textit{Hashed generation:} The main issue with the full generation is that the number of parameters in $W_i$ will increase quadratically between the size of $c_k$ and the filter size and will cause memory issue in training deep adaptive convolution with full generation method. To address this issue, we use a hashing trick \cite{choi2019adaptive} which requires a fraction of important parameters for training. The idea of hash generation is similar to hash embeddings \cite{tito2017hash}, which computes the word embeddings with a weighted sum of \textit{component embeddings} from a shared pool. Similarly to hash embeddings, we generate filter $f_i$ by a weighted sum of \textit{component filters} from a shared pool. The shared pool $E \in \rm {I\!R}^{(l) \times s.d}$ comprises of $l$ number of component filters and is trainable. By using predefined hash functions, we select $z$ component filters from the shared pool and then $f_i$ is generated by a weighted sum of $z$ selected component filters. We explain the generation of $f_i$ step by step as follows:

\begin{enumerate}
    \item By using $z$ different functions $(\mathscr{H}_1, \mathscr{H}_2,..,\mathscr{H}_z)$, we select $z$ \textit{component filters} from a predefined shared pool $E$.
    \item The selected component filters in step 1 are combined as a weighted sum: $f_i = \sum_{j=1}^z p_{i,j} \mathscr{H}_j(\hat{f}_i)$. $(p_{i,1},..,p_{i,z})^\top \in {\rm I\!R}^z$ are known as \textit{importance parameters} for $f_i$ to determine the weight of a linear combination. To ensure that importance parameters are input-specific, we control $p_{i,j}$ as follows:
    \begin{equation}
    p_{i,j} = {w_{i,j}}^\top c_k
\end{equation}
    \noindent where $w_{i,j} \in {\rm I\!R}^g$ is a vector to generate $p_{i,j}$ from a context vector.
\end{enumerate}

\noindent The overall generation of a filter $f_i$ can be denoted in vector notation as follows:
\begin{equation}
    p_{i,j} = (p_{i,1},..,p_{i,z})^\top ,\;
    \mathscr{H} = (\mathscr{H}_1,..,\mathscr{H}_z)^\top ,\;
    f_i = p_{i,j}^\top \mathscr{H}
\end{equation}

 $\mathscr{H}$ represents ID to component filter function for generating component filters from filter IDs $\hat{f}_i$ and is defined as $\mathscr{H}_j = E_{D_j(\hat{f}_i)}$, where $\hat{f}_i$ represents the ID of the filter. $D_j : \{1,..,z\} \rightarrow \{1,..,l\}$ is a hash function, which takes the filter ID as an input and produces a bucket index in $\{1,2,..,l\}$. The row of bucket index in $E$ will be the component filter. The procedure is shown in Figure~\ref{hashgenerationfig}.

Based on the above description, we require the following for each filter generation:

\begin{enumerate}
    \item A trainable shared pool matrix $E \in \rm {I\!R}^{(l) \times s.d}$, where each row represents a component filter.
    \item A weight matrix $W_{i,j} \in {\rm I\!R}^{g\times z}$ to create importance parameters conditioned on inputs.
    \item $z$ different functions $(\mathscr{H}_1, \mathscr{H}_2,..,\mathscr{H}_z)$, each uniformly assigning one of the $l$ component filters to each $f_i$.
\end{enumerate}

The number of parameters required for the generation of each filter will be $z*g$ because we use $z$ number of $p_{i,j}$. We can achieve fairly good performance by choosing a small value of $z$. In this particular setting, we use 5 importance parameters. This helps in a drastic reduction in the number of trainable parameters as compared to the full generation method which requires $s*d*g$ number of parameters for each filter generation. The extra parameters in hash generation come from a shared pool $E$ but its portion is relatively small because the component filters in $E$ are shared across all the filters and its size $l$ can be set to a moderate value (we use $l=20$).

\begin{algorithm}
\SetAlgoLined
\SetKwInOut{Input}{Input}\SetKwInOut{Output}{Output}
\Input{Sentence, predicate, POS tags, dependency tree T.}
\Output{SRL label}
$x \leftarrow  [x^r, x^l, x^p, x^{POS}, x^f, ELMo]$ \\
\For{\upshape{each} epoch}{
$v \leftarrow$ \upshape{BiLSTM encoder}($x$) \\
$H \leftarrow$ \upshape{syntactic encoder}($v$) \\

\For{\upshape{each} $h_i$ \upshape{in} $H$}{

\For{\upshape{For each convolution block in CNNs}}{
$c_i \leftarrow$ \upshape{context vector generation}($h_i$)\\
$\mathscr{F} \leftarrow$ \upshape{filter generation}($c_i$) \\
$O \leftarrow$ \upshape{convolution}($h_i,\mathscr{F}$) \\
$\quad$ \upshape{convolution to $h_i$ with $\mathscr{F}$}\\
$O \leftarrow$ \upshape{max pooling}($O$) \\
$h_i \leftarrow O$\\
}
}
}
\upshape{SRL labels} $\leftarrow$ MLP(softmax($H$)) \\

\caption{\label{algo} Forward propagation of proposed SRL model.}
\end{algorithm}

\subsection{Adaptive convolution}
\label{adaptiveconvolution}
The filter generation network takes an input from the previous convolution block and produces an output except for the first convolution block which takes the output of the syntactic encoder as an input. The input-related generated filters are used by the adaptive convolution to produce output. Specifically, for the $j$-th position of the input window and filter $f_i$, the feature $o_{i,j}$ is computed as follows:
\begin{equation}
o_{i,j} = \phi ({f_i}^\top h_{j:j+s-1} )
\end{equation} 

\noindent where $h_{j:j+s-1}$ is the concatenation of the inputs $[h_j, h_{j+1},...,h_{j+s-1}]$ and $\phi$ is an activation function. By concatenating features $o_{i,j}$ for all the filters in $\mathscr{F}$ for the $j$-th position in the window, a position feature $o_j$ is generated and the output for the adaptive convolution will be $O = [o_1; o_2; ...; o_{m-s+1}]$. After applying max pooling, it will become an input to the next convolution block.  Algorithm \ref{algo} explains the whole procedure.

The output of the convolution operation after max pooling will be ${\rm I\!R}^{n}$ for each $c_k$, where $n$ is the number of filters. The overall output of the adaptive convolution layer will be ${\rm I\!R}^{m \times n}$. 
The output of the adaptive convolution layer is fed into Multi-Layer Perceptron with highway connections followed by a softmax, resulting in an output distribution over $\mathcal{A}$ argument labels for each token in a sentence (${\rm I\!R}^{m \times \mathcal{A}}$). The MLP consists of 10 layers with $ReLU$ activations. To maximize the likelihood of labels, we use categorical cross-entropy as the loss function.

\subsection{Predicate disambiguation:}
Although predicates are already identified for each sentence in CoNLL-2009 dataset, predicate disambiguation is an indispensable task aiming at the identification of a predicate-argument structure in a particular context. This task is comparatively easier to perform, so we use a small portion of the proposed model for this. Given $x_i$ as explained in section \ref{sen-encoder}, we remove Tree LSTM and convolution layer, and use the remaining model for predicate disambiguation.

\section{Experiments}
\begin{table*}
\centering
\caption{\label{conllresults} Results on the CoNLL-2009 English in domain (WSJ) dataset and English out of domain (Brown) datasets.}
\begin{tabular}{lclclcl}
\hline
\multirow{2}{*}{System}                            & \multicolumn{3}{c}{English WSJ}                                                                   & \multicolumn{3}{c}{English Brown}                                                                 \\ \cline{2-7} 
                                                   & $\mathrm{P}$                            & \multicolumn{1}{c}{$\mathrm{R}$}        & $\mathrm{F_1}$                          & \multicolumn{1}{c}{$\mathrm{P}$}        & $\mathrm{R}$                            & \multicolumn{1}{c}{$\mathrm{F_1}$}       \\ \hline
\textit{Local model}                                        &                                &                                &                                 &                                &                                &                                 \\
Zhao \textit{et al.} \cite{zhao2009semantic}          & \_                             & \_                             & 86.2                            & \_                             & \_                             & 74.6                            \\
FitzGerald \textit{et al.} \cite{fitzgerald2015semantic}    & \_                             & \_                             & 86.7                            & \_                             & \_                             & 75.2                            \\
Roth and Lapata \cite{roth2016neural}            & 88.1                           & 85.3                           & 86.7                            & 76.9                           & 73.8                           & 75.3                            \\
Marcheggiani \textit{et al.} \cite{marcheggiani2017simple}    & 88.7                           & 86.8                           & 87.7                            & 79.4                           & 76.2                           & 77.7                            \\
Marcheggiani and Titov \cite{marcheggiani2017encoding}  & 89.1                           & 86.8                           & 88.0                            & 78.5                           & 75.9                           & 77.2                            \\
He \textit{et al.} \cite{he2018syntax}              & 89.7                           & 89.3                           & 89.5                            & 81.9                           & 76.9                           & 79.3                            \\
Cai \textit{et al.} \cite{cai2018full}               & 89.9                           & 89.2                           & 89.6                            & 79.8                           & 78.3                           & 79.0                            \\
Li \textit{et al.} \cite{li2018unified}             & 90.3                           & 89.3                           & 89.8                            & 80.6                           & 79.0                           & 79.8                            \\
Li \textit{et al.} \cite{li2019dependency}          & 89.6                           & 91.2                           & 90.4                            & 81.7                           & 81.4                           & 81.5                            \\
(Ours) SRL-CNN \textit{full}      & 90.5                           & 90.7                           & 90.6                           & 82.1                           & 81.6                           & 81.8                           \\
(Ours) SRL-CNN \textit{hash}      & 90.1                           & 91.0                          & 90.5                           & 82.0                           & 81.5                           & 81.7                            \\
(Ours) SRL-DPCNN \textit{full}    & 90.2                           & \textbf{91.3} & 90.7                           & 82.9 & 81.5                           & 82.2                           \\
(Ours) SRL-DPCNN \textit{hash}    & 91.1                           & 90.2                           & 90.6                           & 82.5                           & 81.3                           & 81.9                            \\
(Ours) SRL-DenseCNN \textit{full} & \textbf{91.2} & 90.6                           & \textbf{90.9} & \textbf{83.1}                           & \textbf{82.6} & \textbf{82.8} \\
(Ours) SRL-DenseCNN \textit{hash} & 90.6                           & 90.8                           & 90.7                            & 82.8                           & 81.1                           & 81.9                           \\ \hline
\textit{\textit{Global model}}             & \multicolumn{1}{l}{}           &                                & \multicolumn{1}{l}{}            &                                & \multicolumn{1}{l}{}           &                                 \\
Bj{\"o}rkelund \textit{et al.} \cite{bjorkelund2010high}        & 88.6                           & 85.2                           & 86.9                            & 77.9                           & 73.6                           & 75.7                            \\
FitzGerald \textit{et al.} \cite{fitzgerald2015semantic}    & \_                             & \_                             & 87.3                            & \_                             & \_                             & 75.2                            \\
Roth and Lapata \cite{roth2016neural}            & 90.0                           & 85.5                           & 87.7                            & 78.6                           & 73.8                           & 76.1                            \\ \hline
\multicolumn{4}{l}{\textit{Ensemble model}}                                                                                                                     &                                & \multicolumn{1}{l}{}           &                                 \\
FitzGerald \textit{et al.} \cite{fitzgerald2015semantic}    & \_                             & \_                             & 87.7                            & \_                             & \_                             & 75.5                            \\
Roth and Lapata \cite{roth2016neural}            & 90.3                           & 85.7                           & 87.9                            & 79.7                           & 73.6                           & 76.5                            \\
Marcheggiani and Titov \cite{marcheggiani2017encoding}  & 90.5                           & 87.7                           & 89.1                            & 80.8                           & 77.1                           & 78.9                            \\ \hline \hline
\end{tabular}

\end{table*}

\begin{table}[]
\centering
\caption{\label{chineseresults} Results on the CoNLL-2009 Chinese test dataset.}
\begin{tabular}[width=\linewidth]{lclc}
\hline
\multirow{2}{*}{System}                             & \multicolumn{3}{c}{Chinese test dataset}                                                          \\ \cline{2-4} 
                                                    & $\mathrm{P}$                            & \multicolumn{1}{c}{$\mathrm{R}$}         & $\mathrm{F_1}$                          \\ \hline
\textit{Local model}                                         &                                &                                 &                                \\
Marcheggiani \textit{et al.} \cite{marcheggiani2017simple}     & 83.4                           & 79.1                            & 81.2                           \\
Marcheggiani and Titov \cite{marcheggiani2017encoding}   & 84.6                           & 80.4                            & 82.5                           \\
He \textit{et al.} \cite{he2018syntax}               & 84.2                           & 81.5                            & 82.8                           \\
Cai \textit{et al.} \cite{cai2018full}                & 84.7                           & 84.0                            & 84.3                           \\
Li \textit{et al.} \cite{li2018unified}              & \textbf{84.8} & 81.2                            & 83.0                           \\
(Ours) SRL-CNN \textit{full}       & 83.4                           & 83.6                            & 83.5                           \\
(Ours) SRL-CNN \textit{hash}       & 83.3                           & 83.1                            & 83.2                           \\
(Ours) SRL-DPCNN \textit{full}     & 84.0                           & 83.6                            & 83.8                           \\
(Ours) SRL-DPCNN \textit{hash}     & 83.9                           & 83.5                            & 83.7                           \\
(Ours) SRL-DenseCNN \textit{full}  & 84.7                           & \textbf{85.1} & \textbf{84.9} \\
(Ours) SRL-DenseCNN \textit{hash}  & 84.5                           & 84.7                            & 84.6                           \\ \hline
\textit{Global model}              & \multicolumn{1}{l}{}           &                                 & \multicolumn{1}{l}{}           \\
Bj{\"o}rkelund \textit{et al.} \cite{bjorkelund2009multilingual} & 84.2                           & \multicolumn{1}{c}{75.1}        & 78.6                           \\
Roth and Lapata \cite{roth2016neural}             & 83.2                           & \multicolumn{1}{c}{75.9}        & 79.4    \\ \hline\hline                      
\end{tabular}

\end{table}

Our model\footnote{The code will be released at \url{https://github.com/kashifmunir92/adaptiveCNN_SRL}} is experimented on CoNLL-2009 dataset for both English and Chinese languages. For English pre-trained embeddings, we use GloVe vectors of 200 dimension \cite{pennington2014glove}. For Chinese pre-trained embeddings, we train a word2vec model (200 dimension) using Wikipedia documents \cite{mikolov2013distributed}. All the other real vectors are randomly initialized with the Guassian distribution with a standard deviation of 0.1. The dimension of lemma embedding is 200, for POS tag embedding is 32 and for predicate identification flag embedding is 16. Additionally, the dimension for ELMo\footnote{For Chinese, we do not use ELMo embeddings as pre-trained ELMo is available for English only.} embedding is 300. To incorporate the dependency information, we use the officially given parses in CoNLL-2009 dataset. For hash convolution, the pool size is 20 and 5 number of \textit{importance parameters}. We optimize the model using Adam optimizer \cite{kingma2014adam} with a learning rate of $1e^{-3}$ and a batch size of 128. For adaptive convolution, we deploy three settings for CNNs as follows:

\noindent $\bullet$ CNN: We use 100 filters with a window size of 3, 100 filters with a window size of 4 and 100 filters with a window size of 5.

\noindent $\bullet$ DPCNN: We use 100 filters for each block of convolution, each having a window size of 3 and depth is kept to 11 for both English and Chinese datasets.

\noindent $\bullet$ DenseCNN: In this scenario, we deploy 75 filters for each convolution block. The inputs to DenseCNN is padded to a fixed length.

\subsection{Results}
\label{results}
% \usepackage{multirow}

%%%%%%%%%%%%%%%%%%%%%%%%%%%%%%%%%%%%%%%%%%%%%%%%%%%%%%%%%%%%%
\begin{table*}
\centering
\caption{\label{conll2008results} Results on the CoNLL-2008 English in domain (WSJ) dataset and English out of domain (Brown) dataset.}
\begin{tabular}{lclclcl}
\hline
\multirow{2}{*}{System}                            & \multicolumn{3}{c}{English WSJ}                                                                   & \multicolumn{3}{c}{English Brown}                                                                 \\ \cline{2-7} 
                                                   & $\mathrm{P}$                            & \multicolumn{1}{c}{$\mathrm{R}$}        & $\mathrm{F_1}$                          & \multicolumn{1}{c}{$\mathrm{P}$}        & $\mathrm{R}$                            & \multicolumn{1}{c}{$\mathrm{F_1}$}       \\ \hline
Johansson and Nugues \cite{johansson2008dependency}          & \_                             & \_                             & 81.7                            & \_                             & \_                             & 69.0                            \\
Zhao \textit{et al.} \cite{zhao2009semantic}    & \_                             & \_                             & 82.1                            & \_                             & \_                             & \_                            \\
Zhao \textit{et al.} \cite{zhao2013integrative}            &\_                           & \_                           & 82.5                            & \_                           & \_                           & \_                            \\
He \textit{et al.} \cite{he2018syntax}    & 83.9                           & 82.7                           & 83.3                            & \_                           & \_                           & \_                            \\
Li \textit{et al.} \cite{li2019dependency}              & 84.5                           & \textbf{86.1}                           & \textbf{85.3}                            & \textbf{74.6}                           & 73.8                           & 74.2                            \\
(Ours) SRL-CNN \textit{full}       & 85.6    & 84.4      & 85.0  & 74.2 & 73.8 &  74.0                \\
(Ours) SRL-CNN \textit{hash}       & 84.5      & 85.3      & 84.9   & 73.9 &73.5 &   73.7                     \\
(Ours) SRL-DPCNN \textit{full}     &  85.0    & 85.4   & 85.2   & 74.3 &73.9 &  74.1                      \\
(Ours) SRL-DPCNN \textit{hash}     & 84.4       & 84.2    & 84.3   & 73.5 & 74.3&  73.9                      \\
(Ours) SRL-DenseCNN \textit{full}  & \textbf{85.8}    & 84.4 & 85.1 & \textbf{74.6} & \textbf{74.8} & \textbf{74.7} \\
(Ours) SRL-DenseCNN \textit{hash}  & 84.8   & 84.6   & 84.7 & 74.0 & 74.4& 74.2                         \\              
 \hline \hline
\end{tabular}

\end{table*}
%%%%%%%%%%%%%%%%%%%%%%%%%%%%%%%%%%%%%%%%%%%%%%%%%%%%%%%%%%%%%

% \usepackage{multirow}
% \usepackage{multirow}

We compare our proposed model with the previously published papers on dependency SRL. Noteworthily, our model performs arguments identification and classification in one shot. Table \ref{conllresults} shows the results of our proposed model for English in domain and out of domain datasets. Our model outperforms most previous approaches of SRL, including ensemble models

For English, SRL-DenseCNN with \textit{full generation} technique of filter generation network gives the best results in terms of $\mathrm{F_1}$ score and precision, while SRL-DPCNN \textit{full} yields the best recall. We outperform syntax-agnostic model of  Li \textit{et al.} \cite{li2019dependency} and syntax aware model of Li \textit{et al.} \cite{li2018unified} by a margin of 0.5\% and $\sim$1.1\% respectively. The same version of the proposed model also performs best on English out of domain dataset, outperforming Li \textit{et al.} \cite{li2018unified} and Li \textit{et al.} \cite{li2019dependency} by a margin of 3\% and 1.3\% respectively, affirming the ability of the proposed model to learn and generalize the latent semantic preferences present in the data.

Table \ref{chineseresults} shows the results of our model on CoNLL-2009 Chinese test dataset. Except for the use of ELMo, we use the same parameters while training the model on Chinese dataset. The results depict that the model overwhelmingly surpasses the previous best performing models, visualizing the proposed model as robust and not sensitive to parameters selection. For Chinese, our model surpasses Li \textit{et al.} \cite{li2018unified} by a margin of 1.9\%. 

\begin{figure}[t]
  \centering
  \begin{minipage}[b]{0.48\textwidth}
    \includegraphics[width=\textwidth]{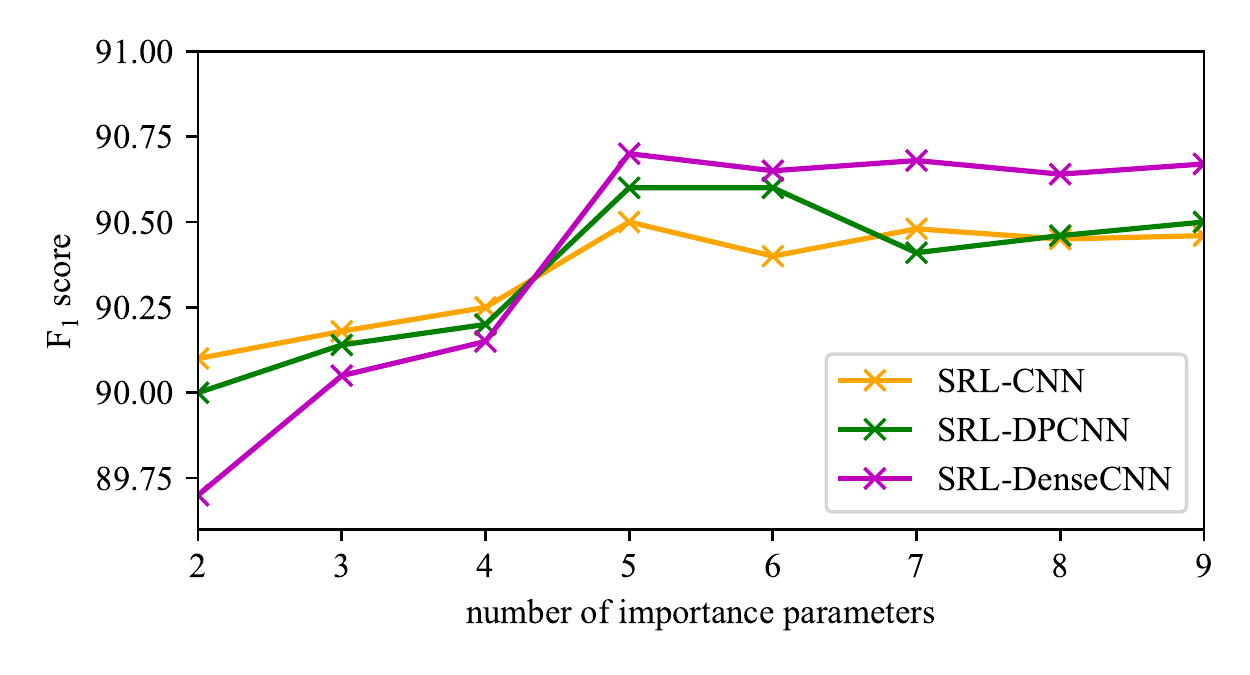}
    \caption{\label{pij}$\mathrm{F_1}$ on English test dataset for different number of importance parameters.}
  \end{minipage}
  \hfill
  \begin{minipage}[b]{0.48\textwidth}
    \includegraphics[width=\textwidth]{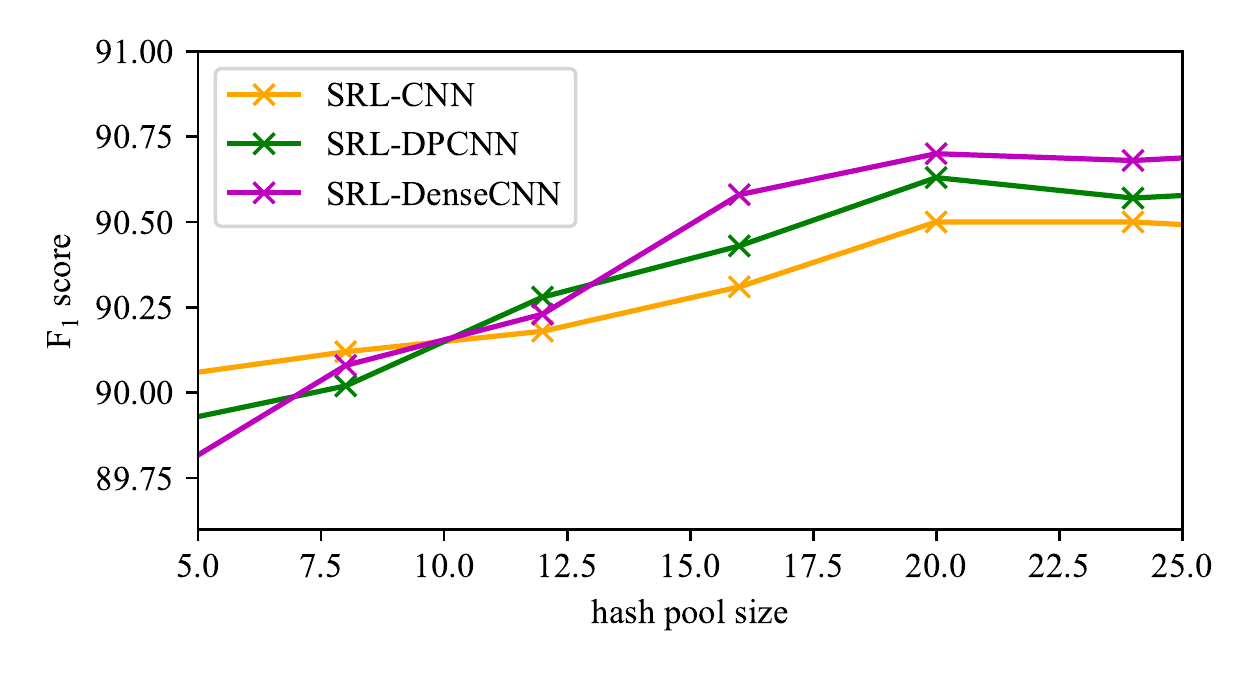}
    \caption{\label{hash}$\mathrm{F_1}$ on English test dataset for different hash pool sizes.}
  \end{minipage}
\end{figure}

We further investigate the effects of hash generation setting on the overall performance of SRL. Figure \ref{pij} and \ref{hash} show how SRL score changes with the number of \textit{importance parameters} and hash pool size. Figures depict that increasing the number of importance parameters and hash pool size does not guarantee a performance boost, however increasing it beyond a certain threshold can affect the model performance. The optimal value of \textit{importance parameter} is 5 and for hash pool size is 20. These findings also confirm that there may exist many redundant parameters in deep neural networks and we can improve the model training speed by selecting important parameters without hurting the model performance.
The difference between full and hash generation techniques is less than 0.3\%. The full generation method performs better than the hash generation method as depicted in the results. But the hash generation method is efficient than the full generation in terms of the model size. Results of the proposed model on both in domain and out of domain datasets show the effectiveness and learning capability of the model.

\subsection{CoNLL-2008 SRL setting}

CoNLL-2009 includes gold predicates beforehand, but predicate identification is an indispensable task for a real-world SRL system. Thus, we use our model for predicates identification and disambiguation as well and evaluate the performance on CoNLL-2008 dataset. Specifically, we use our same model as explained in Section~\ref{methodology} to identify and label predicates. The training scheme remains the same except that we remove the predicate identification flag from the input, while in inference, we perform an additional procedure to identify all the predicates in a given sentence. The arguments labeling is done in a similar way as in CoNLL-2009 setting. 

The overall results on English in-domain (WSJ) and out-of-domain (Brown) test sets are shown in Table~\ref{conll2008results}. On English in-domain test set, our SRL-DPCNN \textit{full} gives $\mathrm{F_1}$ score of 85.2\% which is comparable with the best performing model of Li \textit{et al.} \cite{li2019dependency} (85.3\% $\mathrm{F_1}$). On English out-of-domain test set, our SRL-DenseCNN \textit{full} gives the best performance (74.7\% $\mathrm{F_1}$).

\subsection{Ablation study}
\begin{table}
\centering
\caption{\label{deltafresults} $\Delta \mathrm{F_1}$ represents the performance difference between syntax aware and syntax agnostic models.}
\begin{tabular}{lccc}
\hline
System            & w/o syntax &  with syntax & $\Delta \mathrm{F_1}$ \\
\hline
Marcheggiani and Titov \cite{marcheggiani2017encoding}       & 87.7            & 88.0         & 0.3     \\
He \textit{et al.} \cite{he2018syntax}  & 88.7            & 89.5         & 0.8     \\
Cai \textit{et al.} \cite{cai2018full} & 89.6            & 89.6         & 0       \\
Li \textit{et al.} \cite{li2018unified}  & 88.7            & 89.8         & 1.1     \\
Li \textit{et al.} \cite{li2019dependency}  & 90.4            & 90.4         & 0     \\
\textbf{Ours} SRL-DenseCNN \textit{full}              & 90.0           & 90.9         & 0.9   \\
\hline\hline
\end{tabular}
\end{table}

\begin{table}
\centering
\caption{\label{syntaxencoder} Ablation study to compare the effects of adaptive convolution in SRL system. }
\begin{tabular}{lccc}
\hline
Our system         & $\mathrm{P}$    & $\mathrm{R}$    & $\mathrm{F_1}$    \\
\hline
Ours (syntax-agnostic)  & 90.7 & 89.3 & 90.0 \\
\hline
w/o adaptive convolution       & 90.0 & 87.8 & 88.9 \\
Ours (with GCN) & 90.2 & 88.6 & 89.9 \\
Marcheggiani and Titov \cite{marcheggiani2017encoding} & 89.1 & 86.8 & 88.0 \\
\hline
SRL-DenseCNN \textit{full} &91.2 & 90.6 & 90.9\\
\hline\hline
\end{tabular}
\end{table}

We perform a series of ablation studies on CoNLL-2009 English test dataset to analyze the model.

\begin{table}
\centering
\caption{\label{ablationresults} Ablation study on CoNLL-2009 English test dataset.}
\begin{tabular}{lccc}
\hline
Our system         & $\mathrm{P}$    & $\mathrm{R}$    & $\mathrm{F_1}$    \\
\hline
SRL-DenseCNN \textit{full}  & 91.2 & 90.6 & 90.9 \\
\hline
w/o POS tags       & 91.1 & 89.2 & 90.1 \\
w/o ELMo embedding & 90.5 & 88.9 & 89.7 \\
w/o randomly initialized ($x^r$) & 91.2 & 90.2 & 90.7 \\
w/o pre-trained ($x^p$) & 91.1 & 89.9 & 90.5 \\
w/o lemma ($x^l$) & 90.0 & 91.4 & 90.7 \\
\hline\hline
\end{tabular}
\end{table}

\bigbreak
\noindent \textbf{Word representation:} To interpret the importance of word embedding learned by our model, we carry out experiments with different input settings. Table \ref{ablationresults} shows how our model performs without POS tags information, ELMo embeddings, lemma embeddings ($x^l$), pre-trained embeddings ($x^p$) and randomly initialized embeddings ($x^r$). The effect of POS tags on the overall performance of our model is 0.8\% in terms of $\mathrm{F_1}$ score, which is still better than most of the previous approaches published. The absence of ELMo embedding degrades the performance by 1.2\%. However, the absence of one of the randomly initialized lemma and word embeddings and pre-trained word embeddings has comparatively less impact on the overall performance of our model. These results depict that the presence of these features can enhance the overall SRL performance but our model still provides comparatively better results even in the absence of these features.
 
 \begin{figure}[t]
  \centering
  \begin{minipage}[b]{0.48\textwidth}
    \includegraphics[width=\textwidth]{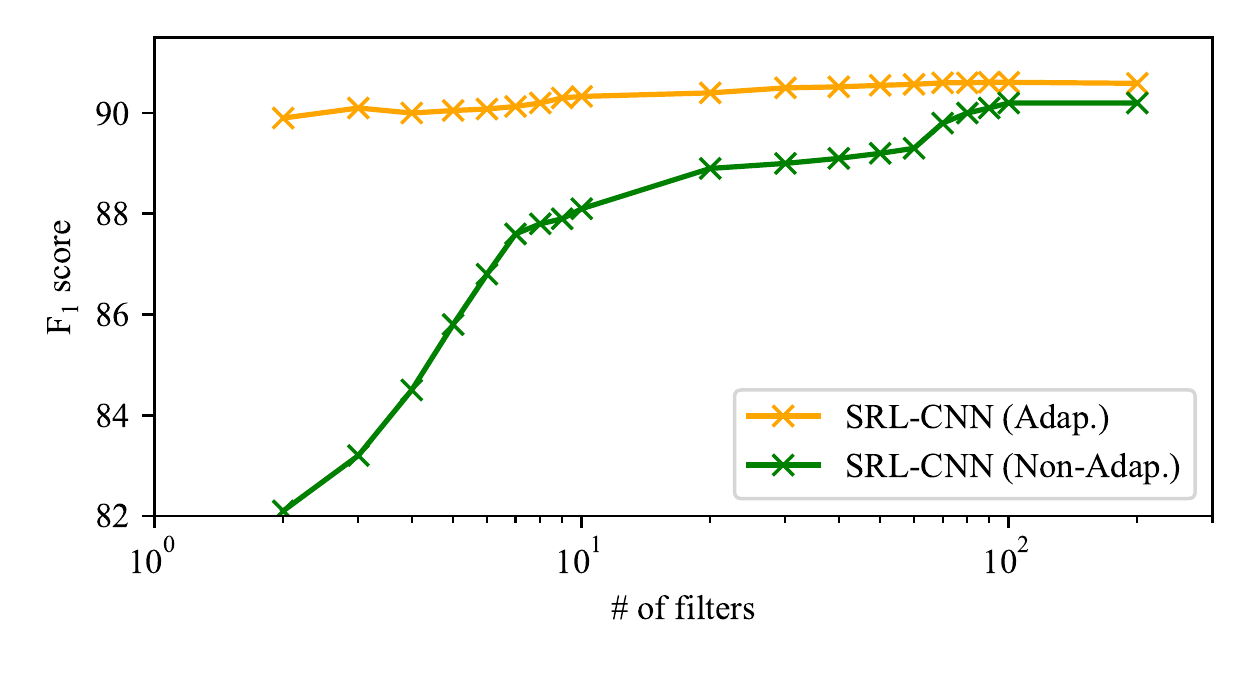}
    \caption{\label{filters}$\mathrm{F_1}$ on English test dataset with different number of filters, x-axis is in logarithmic scale.}
  \end{minipage}
  \hfill
  \begin{minipage}[b]{0.48\textwidth}
    \includegraphics[width=\textwidth]{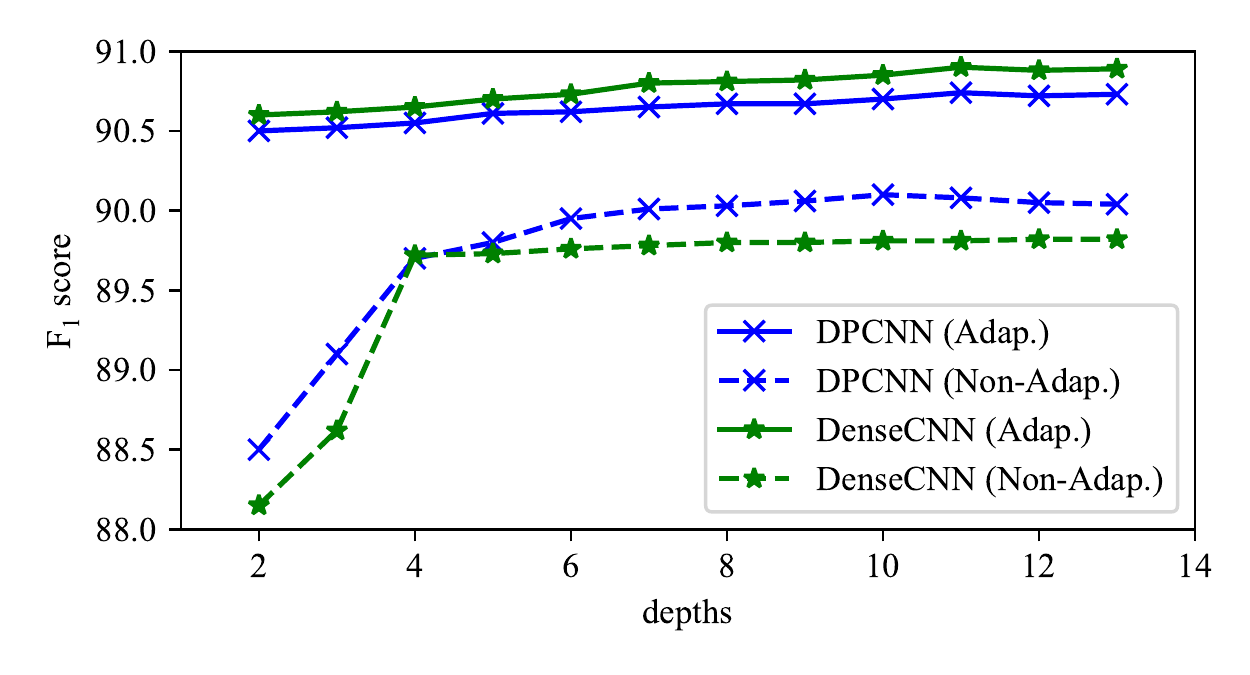}
    \caption{\label{depths}$\mathrm{F_1}$ on English test dataset for different depths.}
  \end{minipage}
\end{figure}
 
 \begin{table*}
\centering
\caption{\label{lasresults} Results on the CoNLL-2009 English test dataset for labeled attachment score ($\mathrm{LAS}$), precision ($\mathrm{P}$), recall ($\mathrm{R}$), semantic labeled score ($\mathrm{F_1}$) and Sem-$\mathrm{F_1}$/ $\mathrm{LAS}$ ratio. We use our SRL-DenseCNN \textit{full} model for comparison.}
\begin{tabular}{lccccc}
\hline
System                                            & $\mathrm{LAS}$   & $\mathrm{P}$             & $\mathrm{R}$             & Sem-$\mathrm{F_1}$        & Sem-$\mathrm{F_1}$/ $\mathrm{LAS}$ \\
\hline
Zhao \textit{et al.} \cite{zhao2009multilingual2} {[}SRL-only{]}                & 86.0  & \_            & \_            & 85.4          & 99.3       \\
Zhao\textit{et al.} \cite{zhao2009multilingual} {[}Joint{]}                   & 89.2  & \_            & \_            & 86.2          & 96.6       \\
\hline
Bj{\"o}rkelund \textit{et al.} \cite{bjorkelund2010high}                          & 89.8  & 87.1          & 84.5          & 85.8          & 95.6       \\
Lei \textit{et al.} \cite{lei2015high}                                 & 90.4  & \_            & \_            & 86.6          & 95.8       \\
Roth and Lapata \cite{roth2016neural}                            & 89.8  & 88.1          & 85.3          & 86.7          & 96.5       \\
Marcheggiani and Titov \cite{marcheggiani2017encoding}                     & 90.34 & 89.1          & 86.8          & 88.0          & 97.41      \\
\hline
He \textit{et al.} \cite{he2018syntax} {[}CoNLL-2009 predicted{]}       & 86.0  & 89.7          & 89.3          & 89.5          & 104.0      \\
He \textit{et al.} \cite{he2018syntax} {[}Gold syntax{]}                & 100   & 91.0          & 89.7          & 90.3          & 90.3       \\
\hline
Li \textit{et al.} \cite{li2018unified} {[}CoNLL-2009 predicted{]}       & 86.0  & 90.5          & 88.5          & 89.5          & 104.7      \\
Li \textit{et al.} \cite{li2018unified} {[}Gold syntax{]}                & 100   & 91.0          & 90.0          & 90.5          & 90.50      \\
\hline
Ours   {[}CoNLL-2009 predicted{]} & 86.0  & 91.2          & 90.6          & 90.9         & 105.6      \\
Ours  {[}CoNLL-2009 Biaffine parser{]} & 90.22  & 91.3          & 90.7          & 91.0         & 100.9      \\
Ours  {[}BIST parser{]} & 90.05  & 91.1          & 90.8          & 90.9         & 101.0      \\
\hline
Ours \textit{full} {[}Gold Syntax{]}          & 100   & \textbf{91.4} & \textbf{91.0} & \textbf{91.2} & 91.2   \\
\hline\hline
\end{tabular}

\end{table*}

 \bigbreak
\noindent \textbf{Adaptive vs non-adaptive convolution:} To verify the effectiveness of adaptive convolution, we compare its performance with varying number of filters (Figure~\ref{filters}) and depths (Figure~\ref{depths}). We further show how the performance varies if we replace adaptive convolution with non-adaptive convolution where filters are not generated dynamically based on the inputs. As can be observed in Figure~\ref{filters}, adaptive CNN provides a performance stability for different number of filters as compared to non-adaptive CNN. Furthermore, the performance of non-adaptive CNN drastically decreases with less number of filters. The performance of non-adaptive CNN with 2 filters is 8.3\% less as compared to that with 100 filters. However, the performance of adaptive CNN with 2 number of filters decreases by 0.7\% only as compared to that with 100 filters.

We see a similar performance variation tendency on depths as can be seen in Figure~\ref{depths}. The performance of adaptive DPCNN  with depth 3 is only 0.6\% less as compared to non-adaptive DPCNN with the best performing depths. This confirms that by using dynamically generated filters, our model can capture the necessary information related to SRL. Adaptive convolution can extract SRL related features with only few filters and shallow depths. This also helps in mitigating the required effort to tune hyperparameters for adaptive convolution.

Table~\ref{parameters} shows that the filter generation by using the full generation method results in a quadratic increase in the number parameters as compared to the hash generation method. The number of parameters in the hash generation method is still larger than that of non-adaptive convolution. But the performance gain of adaptive convolution over non-adaptive convolution is not owing to this increase in the number of parameters. This can be verified by the fact that increasing the number of filters and depth (for non-adaptive convolution) beyond a certain value does not have any impact on model performance. As shown in Figure~\ref{filters}, increasing the number of filters beyond 100 for non-adaptive CNN does not increase performance. Similarly, increasing depth beyond 9 for non-adaptive DenseCNN has no effect on the overall performance (Figure~\ref{depths}). The performance of non-adaptive DPCNN rather decreases when depth is increased beyond 10. This validates that the gain in overall SRL performance is owing to the effectiveness of adaptive convolution, instead of the increased number of parameters. 

\begin{table}
\centering
\caption{\label{parameters} Number of parameters in each model. The input embeddings are not included in the parameters count.}
\begin{tabular}{l|ccc}
\hline
         & CNN    &DPCNN     & DenseCNN     \\

         \hline

Non-adaptive       & 0.6M & 3.5M & 2.7M \\
Hashed generation & 12.9M & 45M & 77M \\
Full Generation & 316.7M & 320.4M & 383.3M \\

\hline\hline
\end{tabular}
\end{table}

\begin{table}
\centering
\caption{\label{LM} Result with different pretrained language models.}
\begin{tabular}{lccc}
\hline
System        & $\mathrm{P}$    & $\mathrm{R}$    & $\mathrm{F_1}$    \\
\hline
SRL-DenseCNN \textit{hash}  & 90.6 & 90.8 & 90.7 \\
\hline
BERT$_{\textrm{BASE}}$       & 91.2 & 90.4 & 90.8 \\
BERT$_{\textrm{LARGE}}$ & 91.1 & 90.9 & 91.0 \\
XLNET$_{\textrm{BASE}}$  & 90.7 & 91.3 & 91.0 \\
XLNET$_{\textrm{LARGE}}$  & 91.5 & 90.9 & 91.2 \\
\hline\hline
\end{tabular}
\end{table}

\bigbreak
\noindent \textbf{Deep encoding effect:} Table \ref{deltafresults} shows the comparison of our SRL-DenseCNN \textit{full} with Marcheggiani and Titov \cite{marcheggiani2017encoding}, He \textit{et al.} \cite{he2018syntax}, Cai \textit{et al.} \cite{cai2018full}, Li \textit{et al.} \cite{li2018unified} and Li \textit{et al.} \cite{li2019dependency}. The reported results are on CoNLL-2009 English test dataset under syntax-aware and syntax agnostic environments. The results show that our model gives better performance improvement with the integration of syntax information as compared to the previous best models.

To further analyze if adaptive convolution helps in the effective encoding of syntactic information, we compare our model with three syntax-aware versions. 
In these experiments, we use the same encoder as Marcheggiani and Titov \cite{marcheggiani2017encoding}. 

\noindent $\bullet$ Our proposed model without adaptive convolution.

\noindent $\bullet$ Replace Tree-LSTM with graph convlotion layer (GCN) as proposed by Marcheggiani and Titov \cite{marcheggiani2017encoding} (i.e. essentially adding adaptive convolution layer above GCN in Marcheggiani and Titov \cite{marcheggiani2017encoding}).

\noindent $\bullet$ Replace Tree-LSTM with graph convlotion layer (GCN) and remove adaptive convolution (i.e. essentially the model of Marcheggiani and Titov \cite{marcheggiani2017encoding}).

The results of these experiments are shown in Table~\ref{syntaxencoder}. As expected, the removal of the adaptive convolution layer results in the performance decline by 2.0\%. However, the inclusion of adaptive convolution above GCN in Marcheggiani and Titov \cite{marcheggiani2017encoding} results in their model's performance improvement by 1.9\%. This performance improvement demonstrates the effectiveness of adaptive convolution which uses dynamically generated filters for capturing information that needs to be disambiguated given the current inputs.

\bigbreak
\noindent \textbf{Syntactic input:} To investigate how the quality of syntactic input affects the SRL performance, we use 4 types of syntactic inputs in our model. 1) predicated parses officially given in CoNLL-2009. 2) Biaffine parser. 3) BIST parser \cite{kiperwasser2016simple}. 4) gold parses given in CoNLL-2009 dataset.

For comparison, we use semantic labeled $\mathrm{F_1}$ as an evaluation metric for SRL performance, labeled attachment score ($\mathrm{LAS}$) to quantify parse quality and Sem-$\mathrm{F_1}$/$\mathrm{LAS}$ as an additional metric for comparison as given by CoNLL-2008 shared task\footnote{CoNLL-2008 task is only for English, while CoNLL-2009 is a multilingual task. The main difference is that predicates are pre-identified for the latter.} \cite{surdeanu2008conll}. 

Table \ref{lasresults} shows that the performance of our model is quite stable with the varying quality of syntactic parses. Secondly, the ratio $\mathrm{F_1}/\mathrm{LAS}$ decreases with the increasing quality of the syntactic parse. Thirdly, when $\mathrm{LAS}$ reaches 100\% for syntactic parse, $\mathrm{F_1}/\mathrm{LAS}$ ratio of our model becomes 91.2\%, advocating the strength of our model. The last conclusion to be drawn from the comparison is that the high quality syntax information can boost SRL performance which lines up with the conclusion drawn by Li \textit{et al.} \cite{li2018unified} and He \textit{et al.} \cite{he2017deep}.

\bigbreak
\noindent \textbf{Replacing ELMo with other language models:} Lastly, we replace ELMo embeddings in the input with other pre-trained language models like BERT \cite{devlin2018bert} and XLNET \cite{yang2019xlnet} to see if our model can achieve further performance improvement. We use our SRL-DenseCNN \textit{hash} model for this experiment on CoNLL-2009 test dataset. The results are shown in Table~\ref{LM}. Both BERT and XLNET help the model to improve the performance over ELMo embeddings for the concerned task. One possible reason behind this improvement is the ability of BERT and XLNET to provide more accurate context information from the text. Furthermore, XLNET embeddings help the model to gain 0.5\% improvement as compared to 0.3\% improvement with BERT embeddings.

\section{Conclusion}
This paper presents a neural framework for semantic role labeling, effectively incorporating a filter generation network to extract important syntactic features encoded by BiLSTM and Tree-LSTM by generating filters conditioned on inputs. The adaptive convolution endows flexibility to existing convolution operations. With the extraction of important syntax information, we are able to enlarge the gap between syntax aware and syntax agnostic SRL systems. We further study a hashing technique which drastically decreases the size of the filter generation network. Lastly, we explore the effects of syntax quality on SRL systems and conclude that the high quality syntax can improve SRL performance. Experiments on CoNLL-2009 dataset validate that our proposed model outperforms most previous SRL systems for both English and Chinese languages.

\bibliographystyle{IEEEtran}
\bibliography{IEEEexample}
\vskip -2\baselineskip plus -1fil
\begin{IEEEbiography}[{\includegraphics[width=1in,height=1.25in,clip,keepaspectratio]{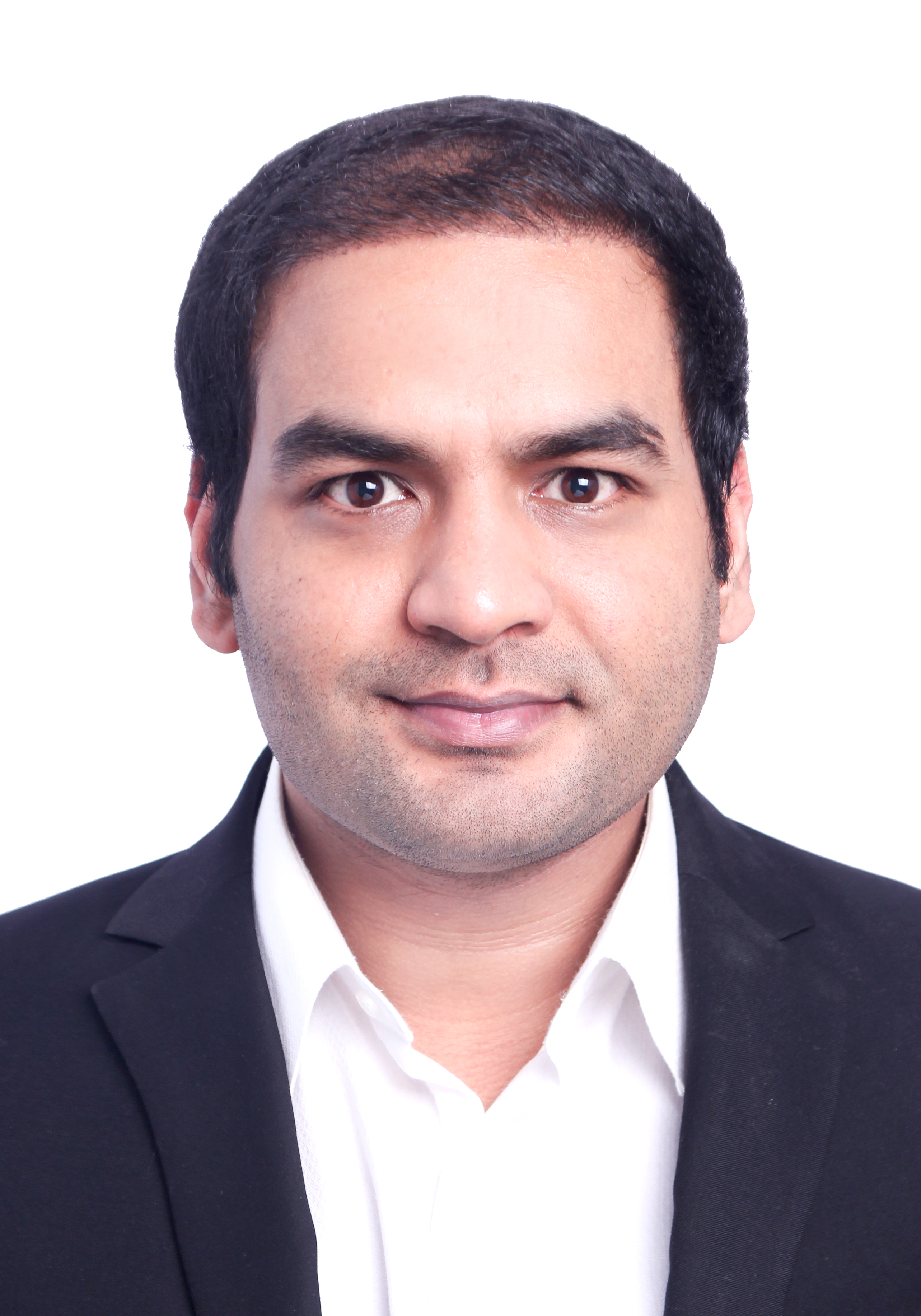}}]{Kashif Munir}
 received Bachelor Engineering degree in electrical engineering from National University of Science and Technology, Islamabad, Pakistan in 2014. Since 2017, he has been a Ph.D. student with the Center for Brain-like Computing and Machine Intelligence of Shanghai Jiao Tong University, Shanghai, China. His research focuses on natural language processing, especially semantic parsing and unsupervised approaches for computing methodologies.
\end{IEEEbiography}
\vskip -2\baselineskip plus -1fil
\begin{IEEEbiography}[{\includegraphics[width=1in,height=1.25in,clip,keepaspectratio]{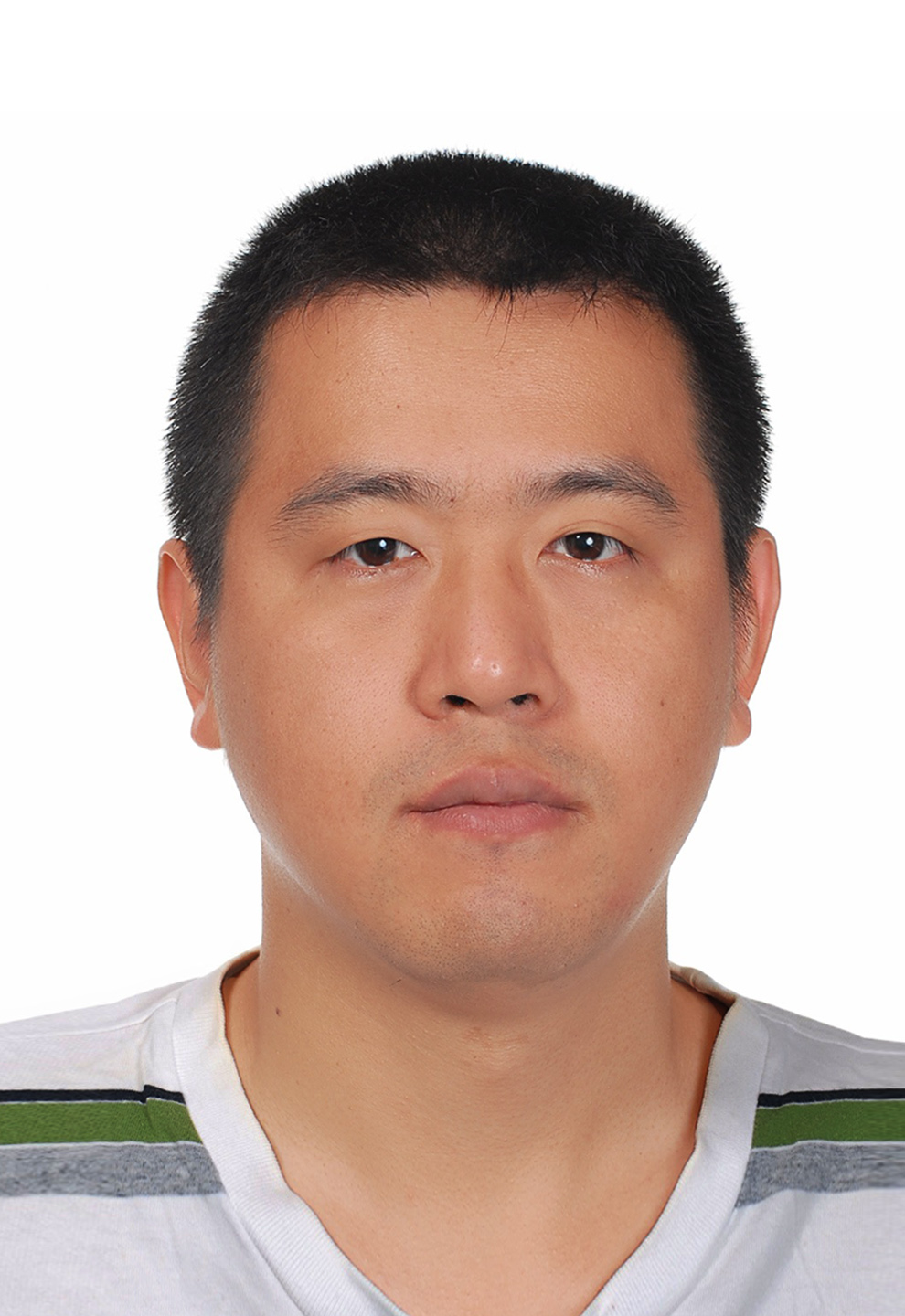}}]{Hai Zhao}
  Hai Zhao received Bachelor Engineering degree in sensor and instrument engineering, and Master Engineering degree in control theory and engineering from Yanshan University in 1999 and 2000, respectively, and PhD in computer science from Shanghai Jiao Tong University, China in 2005. He is currently a full professor at department of computer science and engineering, Shanghai Jiao Tong University. Before he joined the university in 2009, he was a research fellow at City University of Hong Kong from 2006 to 2009. He had been a visiting scholar in Microsoft Research Asia in 2011 and a visiting expert in NICT,Japan in 2012. He is an ACM professional member, and served as (senior) area chair in ACL 2017 on Tagging, Chunking, Syntax and Parsing,  in ACL 2018, 2019 on Phonology, Morphology and Word Segmentation. His research interests include natural language processing and related machine learning, data mining and artificial intelligence.
\end{IEEEbiography}
\vskip -2\baselineskip plus -1fil
\begin{IEEEbiography}[{\includegraphics[width=1in,height=1.25in,clip,keepaspectratio]{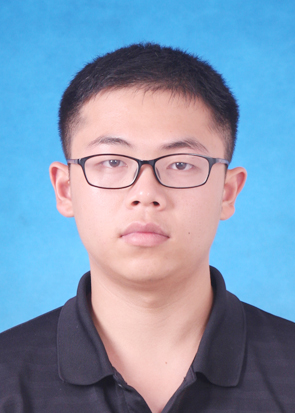}}]{Zuchao Li}
	received the B.S. degree from Wuhan University, Wuhan, China, in 2017. Since 2017, he has been a Ph.D. student with the Center for Brain-like Computing and Machine Intelligence of Shanghai Jiao Tong University, Shanghai, China.  His research focuses on natural language processing, especially syntactic, semantic parsing, and neural machine translation.
\end{IEEEbiography}

\end{document}